\documentclass[letterpaper, 10pt, conference]{ieeeconf}      

\IEEEoverridecommandlockouts                              

\usepackage[T1]{fontenc}
\usepackage{cite}
\usepackage{amsmath,amssymb,amsfonts}
\usepackage{dsfont}
\usepackage[linesnumbered, ruled]{algorithm2e}
\usepackage{graphicx}
\graphicspath{{graphics}}
\usepackage{subcaption}
\usepackage{tabularx}
\usepackage{booktabs}
\usepackage{multirow}
\usepackage{siunitx}
\usepackage{xcolor}
\usepackage[hidelinks]{hyperref}
\hypersetup{
    colorlinks=false,
    linkcolor=blue,
    filecolor=magenta,      
    urlcolor=blue,
}
\usepackage{definitions}
\usepackage[absolute,overlay]{textpos}

\title{\LARGE \bf
Learning Reward Models for Cooperative Trajectory Planning with Inverse Reinforcement Learning and Monte Carlo Tree Search
}

\author{Karl Kurzer$^{1}$, Matthias Bitzer$^{2}$ and J. Marius Z\"ollner$^{3}$
\thanks{$^{1}$Karl Kurzer is with the Institute of Applied Informatics and Formal Description Methods, Karlsruhe Institute of Technology, 76131 Karlsruhe, Germany
        {\tt\small karl.kurzer@kit.edu}}%
\thanks{$^{2}$Matthias Bitzer is with the Bosch Center for Artificial Intelligence, Robert Bosch GmbH, 71272 Renningen, Germany
        {\tt\small matthias.bitzer3@de.bosch.com}}%
\thanks{$^{3}$J. Marius Z\"ollner is with the Institute of Applied Informatics and Formal Description Methods, Karlsruhe Institute of Technology, 76131 Karlsruhe, Germany
        {\tt\small marius.zoellner@kit.edu}}%
}

\begin{document}
	\begin{textblock*}{\textwidth}(19mm,10mm)
		\footnotesize
		\noindent \copyright 2022 IEEE. Personal use of this material is permitted. Permission from IEEE must be obtained for all other uses, in any current or future media, including reprinting/republishing this material for advertising or promotional purposes, creating new collective works, for resale or redistribution to servers or lists, or reuse of any copyrighted component of this work in other works.\\
		\textit{2022 IEEE Intelligent Vehicles Symposium (IV)}
	\end{textblock*}

\maketitle
\thispagestyle{empty}
\pagestyle{empty}

\begin{abstract}
Cooperative trajectory planning methods for automated vehicles can solve traffic scenarios that require a high degree of cooperation between traffic participants. However, for cooperative systems to integrate into human-centered traffic, the automated systems must behave human-like so that humans can anticipate the system's decisions. While Reinforcement Learning has made remarkable progress in solving the decision-making part, it is non-trivial to parameterize a reward model that yields predictable actions. This work employs feature-based Maximum Entropy Inverse Reinforcement Learning combined with Monte Carlo Tree Search to learn reward models that maximize the likelihood of recorded multi-agent cooperative expert trajectories. The evaluation demonstrates that the approach can recover a reasonable reward model that mimics the expert and performs similarly to a manually tuned baseline reward model.
\end{abstract}

\section{INTRODUCTION}
Reinforcement Learning (RL) based approaches frequently make use of manually specified reward models \cite{Wolf2018, Kurzer2021}. In environments where systems need to interact with humans, their decisions must be comprehensible and predictable. As the complexity of the reward model rises, the manual parametrization of the same to generate the desired behavior becomes quickly infeasible.
In the case of driving, it is clear that various features influence the reward of any given trajectory \cite{Naumann2020}. While tuning the weighting of features to create the desired behavior in a diverse set of scenarios is tedious and error-prone, Inverse Reinforcement Learning (IRL) has proven to be able to recover the underlying reward model from recorded trajectories that demonstrate expert behavior in areas such as robotics and automated driving \cite{Ng2000, Abbeel2004, Ziebart2008, Kuderer2015, Wulfmeier2016b}.

\begin{figure}
	\centering
	\def\svgwidth{\columnwidth}
	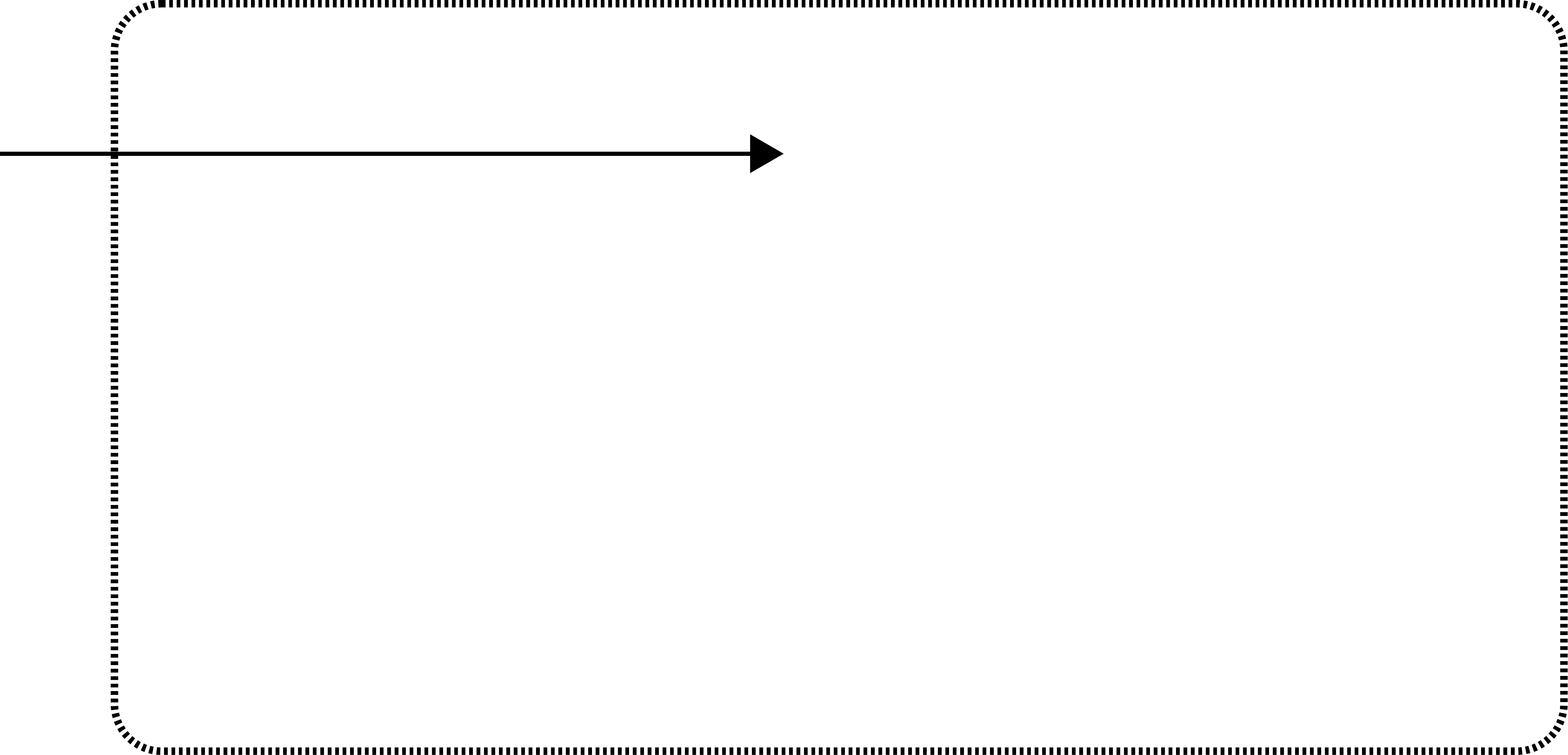
	\caption{Overview of the System: At first, an initial set of expert trajectories $\trajectorySpace_\expert$ is generated. Then the cooperative trajectory planning algorithm computes a set of sample trajectories $\trajectorySpace_\sample$ using the randomly initialized reward model. Next, using the $\trajectorySpace_\expert$ and $\trajectorySpace_\sample$, the likelihood of the parameters $\weights$ given the expert trajectories is increased using gradient ascent. Finally, the process repeats with the cooperative trajectory planning algorithm, sampling new trajectory samples until convergence}
	\label{fig:overview}
\end{figure}

This work builds on an existing cooperative trajectory planning algorithm \cite{Kurzer2018b} to generate expert trajectories. Its contribution is twofold. The first is a system that conducts Guided Cost Learning (GCL), a sampling-based Maximum Entropy Inverse Reinforcement Learning method with Monte Carlo Tree Search (MCTS) to efficiently solve the forward RL problem in a cooperative multi-agent setting. The second is an evaluation that compares a linear and nonlinear reward model to a manually designed one. It is shown that the performance of the learned models is similar to or better than the tediously tuned baseline. An overview of the system is depicted in Fig.~\ref{fig:overview}.
\section{RELATED WORK}
While the task in RL is the deduction of an optimal policy from interactions of an agent with the environment based on a reward model \cite{Sutton2018}, see \eqref{eq:optimal_policy}, the opposite is the case for IRL \cite{Arora2021}.
Here, the task is to infer the underlying reward model that the optimal policy aims to maximize \cite{Ng2000}.
Since the reward model is the most succinct and transferable description of an agent's behavior \cite{Abbeel2004}, a close approximation of the underlying reward model will yield a behavior that is similar to the behavior that results from the optimal policy, i.e., the expert behavior.

Early work in IRL performed feature matching rather than estimating the true underlying reward function \cite{Abbeel2004} to learn driving styles in a discrete driving simulator. More recently, driving styles are learned using continuous trajectories, and action spaces \cite{Kuderer2015}, including additional features that impact driver preference \cite{Naumann2020}. 

Wulfmeier et al. demonstrate the effectiveness of learning nonlinear reward models building on Maximum Entropy IRL \cite{Ziebart2008} using Deep Neural Networks \cite{Wulfmeier2016a}, which they extended to learning cost maps for path planning from raw sensor measurements \cite{Wulfmeier2016b}.

Further improvements in the approximation of the partition function and the efficiency of IRL in combination with RL have been proposed by Guided Cost Learning \cite{Finn2016}. By adapting the IRL procedure, the method yields both a cost function and policy given expert demonstrations using sampling-based methods. In addition, even more efficient one-shot sampling methods have been proposed \cite{Wu2020}.

In contrast, the following work learns a linear and a nonlinear reward model that are integrated into a cooperative multi-agent trajectory planning algorithm using a continuous action and state space.
\section{PROBLEM STATEMENT}
The policy
\begin{equation}
	\label{eq:policy}
	\policy = \policyFull \qquad \action \in \actionSpace(s)
\end{equation}
is a probability distribution over actions conditioned on states.
The trajectory
\begin{equation}
	\label{eq:trajectory}
	\trajectory = {(\trajectoryFull)} 
\end{equation}
defines a path through an MDP.
Using these definitions a policy over trajectories
\begin{equation}
	\label{eq:policy_trajectories}
	\policyTrajectory(\trajectory) = \policyTrajectory(\trajectoryFull) = \prod_{t=0}^{T-1} \policyFullT
\end{equation}
can be defined assuming a deterministic start state distribution and transition model.
The return
\begin{equation}
	\label{eq:return}
	\return_{\weights}(\trajectory) = \sum_{(\stateT,\actionT) \in \trajectory}^{T}{\discountfactor^{t}\rewardmodel_{\weights}(\stateT, \actionT)} 
\end{equation}
of a trajectory $\trajectory$ equals its accumulated discounted (\discountfactor{}) reward at time step $t$, taking action $\actionT$ in state $\stateT$ \cite{Sutton2018}.
The value function of a policy $\policy$ for an MDP with a reward model parameterized by $\weights$ is the expectation of the return of trajectories sampled from that policy,
\begin{equation}\label{eq:value_function}
	V^{\policy}_{\weights}(\state) = \expectation_{\trajectory \sim \policyTrajectory}\left[\return_{\weights}(\trajectory)\right].
\end{equation}
While the forward RL problem is solved by finding the optimal policy
\begin{equation}\label{eq:optimal_policy}
	\policyOpt = \amax{\policy}V^{\policy}_{\weights}(\state),
\end{equation}
the inverse RL problem is solved by finding the parameters \weights{} so that,
\begin{equation}\label{eq:optimal_parameters}
	V^{\policyE}_{\weights}(\state) \geq V^{\policy}_{\weights}(\state) \quad \forall \policy \in \policySpace,
\end{equation}
with \policyE{} being the expert policy as part of the policy space $\policySpace$.

Please refer to \cite{Sutton2018} and \cite{Arora2021} for an in-depth introduction to RL and IRL.

This work aims to learn the parameters \weights{} of a reward model for cooperative trajectory planning so that the optimal trajectories of the planning algorithm are similar to the demonstrated expert trajectories. That means that the expert policy yields the highest state value of all policies given the parametrization of the reward model \eqref{eq:optimal_parameters}.

\section{APPROACH}
To learn a reward model from expert demonstrations so that the behavior sampled from the optimal policy based on this reward model resembles the expert demonstrations, IRL is used. Concisely, this work makes use of a cooperative trajectory planning algorithm based on MCTS \cite{Kurzer2018b}, and Maximum Entropy Inverse Reinforcement Learning \cite{Ziebart2008}, yielding a system that is similar to Guided Cost Learning \cite{Finn2016}.

The MCTS is used to solve the (forward) RL problem, i.e. finding the optimal policy/action given a reward model and generating near optimal trajectory samples $\trajectorySpace_\sample$ for that policy. Using these trajectories in combination with the expert trajectories $\trajectorySpace_\expert$ Maximum Entropy Inverse Reinforcement Learning is used to conduct a gradient ascent step increasing the likelihood of the parameters $\weights$ given the expert trajectories, \eqref{eq:irl_max_log_likelihood}, \eqref{eq:irl_gradient_ascent_log_likelihood}, see Fig.~\ref{fig:overview}.

While the trajectory planning algorithm explicitly encodes interaction between agents, the IRL procedure treats the resulting trajectories as if they would stem from a single agent MDP, with other agents being part of the environment. On the one hand, this has the advantage that the resulting reward model is more robust towards changes in the number of agents. However, on the other hand, it has the drawback of a non-stationary environment, as changes in the reward model (after each gradient ascent step of the IRL procedure) change the behavior of all agents, potentially destabilizing the training process \cite{Mnih2015}. Nevertheless, this was not found to be the case in this work.

\subsection{Solving the Forward RL Problem}
Most IRL algorithms require a method that evaluates the current parameters of the reward model within the algorithm (i.e., finding an optimal policy given the current reward model). For complex tasks where finding a solution to the forward RL problem is hard, IRL can quickly become impractical \cite{Finn2016, Wu2020}. Moreover, the task of finding an optimal policy for any given MDP is usually much harder than finding an optimal action (or trajectory) for the same MDP given a specific state (learning vs. planning)\cite{Sutton2018}. The MCTS-based cooperative trajectory planning algorithm \cite{Kurzer2018b} is thus vital in the IRL setting, as it generates near-optimal trajectories for arbitrary reward models quickly. Thus, this work employs said algorithm to solve the forward RL problem.

\subsection{Reward Model}
The reward model is a central part of an RL system, as the goal of RL is to maximize the cumulative discounted reward by finding the optimal policy \cite{Sutton2018}.

Initially IRL applied solely linear reward models, that are represented as a linear combination of features \features{} and parameters \weights{} \eqref{eq:linear_reward_model}\cite{Abbeel2004}. However, especially for larger RL problems, linear reward models have been outperformed by nonlinear reward models such as neural networks \cite{Wulfmeier2016a, Finn2016}.
This work uses both a linear reward model and a nonlinear reward model in the form of a neural network.

\subsubsection{Features}
Similar to many other planning methods, the cooperative trajectory planner assumes the desired lane \desiredLane{} as well as the desired velocity \desiredVelocity{} for each agent \cite{Naumann2020}. State and action dependent features $\features{}$ are scalar values that consider specific characteristics of a state and action. Each feature is evaluated for each time step $t$ of the trajectory.

\begin{equation}\label{eq:feature_trajectory}
	\feature(\trajectory)=\frac{1}{T}\sum_{(\stateT,\actionT) \in \trajectory}^{T}\feature(\stateT,\actionT).
\end{equation}The parameters \weights{} are identical for all agents, however, features are not. Therefore, all features are normalized to lie between $[-1, 1]$ for the length $T$ of a trajectory \trajectory{}. The feature for the desired lane is defined as
\begin{equation}\label{eq:feature_desiredLane}
	\feature_\mathrm{desLane}(\trajectory)=\frac{1}{T}\sum_{(\stateT,\actionT) \in \trajectory}^{T} \max\left(1-\left|\lane_t-\desiredLane\right|,-1\right),
\end{equation} encouraging the agent to drive on the desired lane.
A deviation from the desired velocity \desiredVelocity{} larger than \SI{10}{\percent} results in a negative feature value, 
\begin{equation}\label{eq:feature_desiredVelocity}
	\feature_\mathrm{desVelocity}(\trajectory)=\frac{1}{T}\sum_{(\stateT,\actionT) \in \trajectory}^{T} \max\left(1-10\left|\frac{v_t}{\desiredVelocity}-1\right|,-1\right).
\end{equation}
Similarly, deviating more than a quarter of the lane width \laneWidth{} from the lane center \laneCenter{} yields a negative feature value,
\begin{equation}\label{eq:feature_laneCenter}
	\feature_\mathrm{laneCenter}(\trajectory)=\frac{1}{T}\sum_{(\stateT,\actionT) \in \trajectory}^{T} \max \left(1-\frac{\left|\laneCenter-y_t\right|}{\laneWidth/4},-1\right).
\end{equation}
A proxy value for the acceleration $a$ of an action is determined to avoid excessive accelerations,
\begin{equation}\label{eq:cost_acceleration}
	\cost_\mathrm{acceleration}=\frac{1}{\gravity}\sqrt{\frac{\int_{t}^{t+\Delta T}(a(t))^{2} \dif t}{\Delta T}.}
\end{equation}
If this value is larger than an eighth of the gravity \gravity{}, the feature turns negative, 
\begin{equation}\label{eq:feature_acceleration}
	\feature_\mathrm{acceleration}(\trajectory)=\frac{1}{T}\sum_{(\stateT,\actionT) \in \trajectory}^{T} \max \left(1-\frac{\cost_\mathrm{acceleration}}{\gravity/8},-1\right).
\end{equation}
In addition, the following binary features are defined for trajectories that either result in collisions \eqref{eq:feature_collision}, invalid states (i.e. an agent drives off the road) \eqref{eq:feature_invalid_state} or invalid actions \eqref{eq:feature_invalid_action} (i.e. an agent executes a physically impossible action). Each of these binary features mark a terminal state.
\begin{equation}\label{eq:feature_collision}
	\feature_\mathrm{collision}(\trajectory) = 
	\begin{cases}
		1~&{\text{ if }}~\trajectory\in {\{\mathrm{collision}\}}~\\0~&{\text{ if }}~\trajectory\notin {\{\mathrm{collision}\}}
	\end{cases}
\end{equation}

\begin{equation}\label{eq:feature_invalid_state}
	\feature_\mathrm{invalid\ state}(\trajectory) = 
	\begin{cases}
		1~&{\text{ if }}~\trajectory\in {\{\mathrm{invalid\ state}\}}~\\0~&{\text{ if }}~\trajectory\notin {\{\mathrm{invalid\ state}\}}
	\end{cases}
\end{equation}

\begin{equation}\label{eq:feature_invalid_action}
	\feature_\mathrm{invalid\ action}(\trajectory) = 
	\begin{cases}
		1~&{\text{ if }}~\trajectory\in {\{\mathrm{invalid\ action}\}}~\\0~&{\text{ if }}~\trajectory\notin {\{\mathrm{invalid\ action}\}}
	\end{cases}
\end{equation}

\subsubsection{Linear Reward Model}
The linear reward model is a linear combination of the parameters \weights{} and their respective features \features{}, 
\begin{equation}\label{eq:linear_reward_model}
	\rewardmodel_{\weights}(\stateT, \actionT)= \weightsT \featuresT.
\end{equation}
The feature count is normalized using the length of the trajectory. Since each feature is bounded between $[-1, 1]$ the return of a trajectory is bounded between $[-\lvert\lvert\weights\rvert\rvert, \lvert\lvert\weights\rvert\rvert]$.
\subsubsection{Nonlinear Reward Model}
To allow for a more complex reward structure, a nonlinear reward model in the form of a fully connected neural network is introduced,
\begin{equation}\label{eq:nonlinear_reward_model}
	\rewardmodel_{\weights}(\stateT, \actionT, \state_{t-1})=W_{2}\Gamma(W_{1} \feature(\stateT, \actionT, \state_{t-1})).
\end{equation}
It consists of two layers, with parameters $W_{1}$ and $W_{2}$, respectively. The first layer is followed by a ReLU activation function $\Gamma$. The input to the network are the features for the linear model in addition to the values of $\feature_\mathrm{desLane}$, $\feature_\mathrm{desVelocity}$ and $\feature_\mathrm{laneCenter}$ at the previous time step.

\subsection{Maximum Entropy Inverse Reinforcement Learning}
IRL learns the parameters \weights{} of a parameterized reward model $\rewardmodel_{\weights}$ so that the expert policy \policyE{} becomes the optimal policy given the reward model \cite{Ng2000}.

Instead of requiring access to the expert policy \policyE{} itself, it is sufficient to observe trajectories $\trajectorySpace_\expert$ that originate from that policy \cite{Abbeel2004}.
\begin{equation}
	\trajectoryE = (\trajectoryFull) \quad \actionT \sim \policyE(\actionT|\stateT, \weights)
\end{equation}
Similarly to the policy \policy{} as a distribution over actions \eqref{eq:policy}, a policy \policyTrajectory{} as a distribution over trajectories can be defined \eqref{eq:policy_trajectories}.

A prominent method for IRL is Maximum Entropy Inverse Reinforcement Learning \cite{Ziebart2008}, which assumes a probabilistic model for expert behavior.
Using the definition of a policy over trajectories \eqref{eq:policy_trajectories} Maximum Entropy IRL specifies the distribution over expert trajectories conditioned on the parameters of the reward model
\begin{equation}
	\label{eq:irl_policy_expert}
	\policyTrajectory_\expert(\trajectory)=\frac{e^{\return_{\weights}(\trajectory)}}{\partitionF}.
\end{equation}
This model implies that the probability of an expert trajectory increases exponentially with its return.
With the numerator being the exponentiated return of a trajectory \eqref{eq:return} and the denominator the partition function \eqref{eq:irl_partition_function}, the integral of the exponentiated return of all trajectories.
\begin{equation}
	\label{eq:irl_partition_function}
	\partitionF = \int e^{\return_{\weights}(\trajectory)} \dif\trajectory
\end{equation}
The likelihood of the parameters \weights{} given the expert trajectories $ \trajectorySpace_\expert$ is defined with
\begin{equation}
	\label{eq:irl_likelihood}
	\likelihood(\weights | \trajectorySpace_\expert) = \prod_{\trajectory \in \trajectorySpace_\expert} \policyTrajectoryFull.
\end{equation}Applying the logarithm to \eqref{eq:irl_likelihood}, yields the Log-likelihood
\begin{equation}
	\label{eq:irl_log_likelihood} \logLikelihood(\weights | \trajectorySpace_\expert) = \sum_{\trajectory \in \trajectorySpace_\expert} \left(\return_{\weights}(\trajectory) - \log{\partitionF}\right),
\end{equation}which is proportional to
\begin{equation}
	\label{eq:irl_likelihood_proportional} \frac{1}{|\trajectorySpace_\expert|}\sum_{\trajectory \in \trajectorySpace_\expert} \return_{\weights}(\trajectory) - \log{\partitionF}.
\end{equation}
The maximization of the Log-likelihood\footnote{in the following, the Log-likelihood refers to the proportional Log-likelihood} \eqref{eq:irl_likelihood_proportional} through the parameters \weights{} will result in the parameters that best explain the expert trajectories.
\begin{equation}
	\label{eq:irl_max_log_likelihood}
	\max_{\weights \in \weightsSpace}\frac{1}{|\trajectorySpace_\expert|}\sum_{\trajectory \in \trajectorySpace_\expert} \return_{\weights}(\trajectory) - \log{\partitionF}
\end{equation}
Using the gradient of the Log-likelihood in a gradient ascent step, locally optimal parameters can be found \eqref{eq:irl_gradient_ascent_log_likelihood}.
\begin{equation}
	\label{eq:irl_gradient_ascent_log_likelihood}
	\weights \gets \weights + \learningrate \nabla_{\weights} \logLikelihood(\weights)
\end{equation}
Given the equalities $\frac{d}{d x}\ln{f(x)} = \frac{f^{\prime}(x)}{f(x)}$\footnote{\href{https://en.wikipedia.org/wiki/Logarithmic_derivative}{Logarithmic derivative}} and $\frac{d}{d x} e^{f(x)} = f^{\prime}(x) e^{f}$\footnote{\href{https://en.wikipedia.org/wiki/Exponential_function}{Exponential derivative}} the gradient of the Log-Likelihood \eqref{eq:irl_gradient_log_likelihood_derivation} can be formulated as an expectation \eqref{eq:irl_gradient_log_likelihood}.
\begin{equation}
	\label{eq:irl_gradient_log_likelihood_derivation}
	\begin{split}
		\nabla_{\weights} \logLikelihood(\weights) & = \gradientReturnE - \nabla_{\weights}\log{\partitionF} \\
		& = \gradientReturnE - \frac{ \nabla_{\weights}\partitionF}{\partitionF} \\
		& = \gradientReturnE - \frac{ \nabla_{\weights}\int e^{\return_{\weights}(\trajectory)} \dif\trajectory}{\partitionF} \\
		& = \gradientReturnE - \frac{\int e^{\return_{\weights}(\trajectory)}}{\partitionF} \gradientReturn \dif\trajectory \\
		& = \gradientReturnE - \int \policyTrajectory_\expert(\trajectory) \gradientReturn \dif\trajectory
	\end{split}
\end{equation}
\begin{equation}
	\label{eq:irl_gradient_log_likelihood}
	\nabla_{\weights} \logLikelihood(\weights) = \gradientReturnE - \expectation_{\trajectory \sim \policyTrajectory_\expert(\trajectory)}\left[\gradientReturn\right]
\end{equation}
\subsection{Guided Cost Learning}
Guided Cost Learning (GCL) is an algorithm that combines sampling-based Maximum Entropy IRL with RL \cite{Finn2016}.

Since the partition function \eqref{eq:irl_partition_function} can only be calculated for small and discrete MDPs, it cannot be computed for the cooperative trajectory planning problem. GCL circumvents this problem by sampling to approximate it.

It estimates the partition function \eqref{eq:irl_partition_function} using the distribution over trajectories generated by a sampling-based method (in this work, the MCTS-based cooperative trajectory planner \cite{Kurzer2018b}) \eqref{eq:irl_policy_trajectories_mcts}). With the ideal proposal density for importance sampling being the distribution that yields the lowest variance \cite{Finn2016}
\begin{equation}\label{eq:irl_optimal_proposal}
	\policyTrajectory_\sample^\ast(\trajectory) \propto e^{\return_{\weights}(\trajectory)}.
\end{equation}
The key concept of GCL is the adjustment of this sampling distribution to the distribution that follows from the current reward model \eqref{eq:irl_policy_expert}. In order to achieve this within the MCTS, this work introduces a probabilistic final selection policy named \textit{Softmax Q-Proposal},
\begin{equation}\label{eq:irl_softmax_q}
	\policy[MCTS](\action | \stateZero) = \frac{e^{cQ^{\policy}(\stateZero, \action)}}{\sum_{\action \in \actionSpace(\stateZero)} e^{cQ^{\policy}(\stateZero, \action)}}.
\end{equation}
The numerator is the exponentiated state-action value $Q^{\policy}(\stateZero, \action)$ (i.e., the expected return \eqref{eq:return}) of taking action \action{} in root state $\stateZero$ over the sum of the state-action values of all explored actions $\actionSpace$ in the root state $\stateZero$. The coefficient $c$ can be used to scale the variance of the distribution, its value is determined empirically.
Based on \eqref{eq:policy_trajectories} this results in the following distribution over trajectories
\begin{equation}
	\label{eq:irl_policy_trajectories_mcts}
	\policyTrajectory(\trajectory) =  \policyTrajectory_\mathrm{MCTS}(\trajectoryFull) = \prod_{t=0}^{T-1} \policy[MCTS](\actionT|\stateT).
\end{equation}
Applying importance sampling (see \ref{sec:preliminiaries_importance_sampling}) the expectation in \eqref{eq:irl_gradient_log_likelihood} can be calculated using the policy $\policyTrajectory_\sample(\trajectory)$ \eqref{eq:irl_expectation_gradient_return_derivation},
\begin{equation}
	\label{eq:irl_expectation_gradient_return}
	\expectation_{\trajectory \sim \policyTrajectory_\expert(\trajectory)}\left[\gradientReturn\right] = \expectation_{\trajectory \sim \policyTrajectory_\sample(\trajectory)}\left[\frac{e^{\return_{\weights}(\trajectory)}}{\policyTrajectory_\sample(\trajectory)\partitionF}\gradientReturn\right]
\end{equation}
Further the partition function can be approximated using importance sampling as well \eqref{eq:irl_partition_function_expectation}, 
\begin{equation}
	\label{eq:irl_partition_function_approximation}
	\partitionFApprox \defined \frac{1}{|\trajectorySpace_\sample|} \sum_{\trajectory \in \trajectorySpace_\sample} \frac{e^{\return_{\weights}(\trajectory)}}{\policyTrajectory_\sample(\trajectory)}.
\end{equation}
Substituting the expectation in \eqref{eq:irl_gradient_log_likelihood} with \eqref{eq:irl_expectation_gradient_return} as well as the partition function \eqref{eq:irl_partition_function} with \eqref{eq:irl_partition_function_approximation}, the final approximation of the gradient can be obtained \eqref{eq:irl_gradient_log_likelihood_substitution}.
\begin{equation}
	\begin{split}
		\label{eq:irl_gradient_log_likelihood_substitution}
		\nabla_{\weights} \logLikelihood(\weights) & = \gradientReturnE - \expectation_{\trajectory \sim \policyTrajectory_\expert(\trajectory)}\left[\gradientReturn\right]\\
		& = \gradientReturnE - \expectation_{\trajectory \sim \policyTrajectory_\sample(\trajectory)}\left[\frac{e^{\return_{\weights}(\trajectory)}}{\policyTrajectory_\sample(\trajectory)\partitionF}\gradientReturn\right]\\
		& \approx  \gradientReturnE - \frac{1}{|\trajectorySpace_\sample|} \sum_{\trajectory \in \trajectorySpace_\sample}\frac{e^{\return_{\weights}(\trajectory)}}{\policyTrajectory_\sample(\trajectory)\partitionFApprox}\gradientReturn
	\end{split}
\end{equation}
Given this form of the gradient, the proposed Softmax Q-IRL algorithm (Alg.~\ref{alg:irl_softmax_q}) performs gradient ascent, converging towards the expert behavior.

The necessary data sampling routine (Alg.~\ref{alg:irl_softmax_q} Line~\ref{alg:generate_samples}) is depicted in Alg.~\ref{alg:irl_data_sampling}. It generates the sample trajectories $\trajectorySpace_\sample$ as well as their policies $\policySpace$. Here, \agentSpace{} denotes the number of agents in the respective scenario.

\section{Experiments}
For each scenario, a set $\trajectorySpace_\expert$ of 50 expert trajectories is generated that depicts (approximately) optimal behavior using the cooperative trajectory planner\footnote{\url{https://url.kurzer.de/ProSeCo}}. Each trajectory has a length of \SI{10.4}{\second}. While a scenario has a fixed number of agents and obstacles, its start state is sampled from a distribution. Explicitly, the longitudinal and lateral positions of the agents are sampled from a normal distribution. Further, different random seeds are used to initialize the sampling-based trajectory planning algorithm.

The linear and the nonlinear reward models were trained for 2000 gradient steps with a learning rate of 0.0005.

The source code of this work is available online\footnote{\url{https://url.kurzer.de/ProSeCo-IRL}}.

\begin{figure}
	\footnotesize
	\centering
	\begin{subfigure}[b]{0.49\columnwidth}
		\centering
		\def\svgwidth{\columnwidth}
		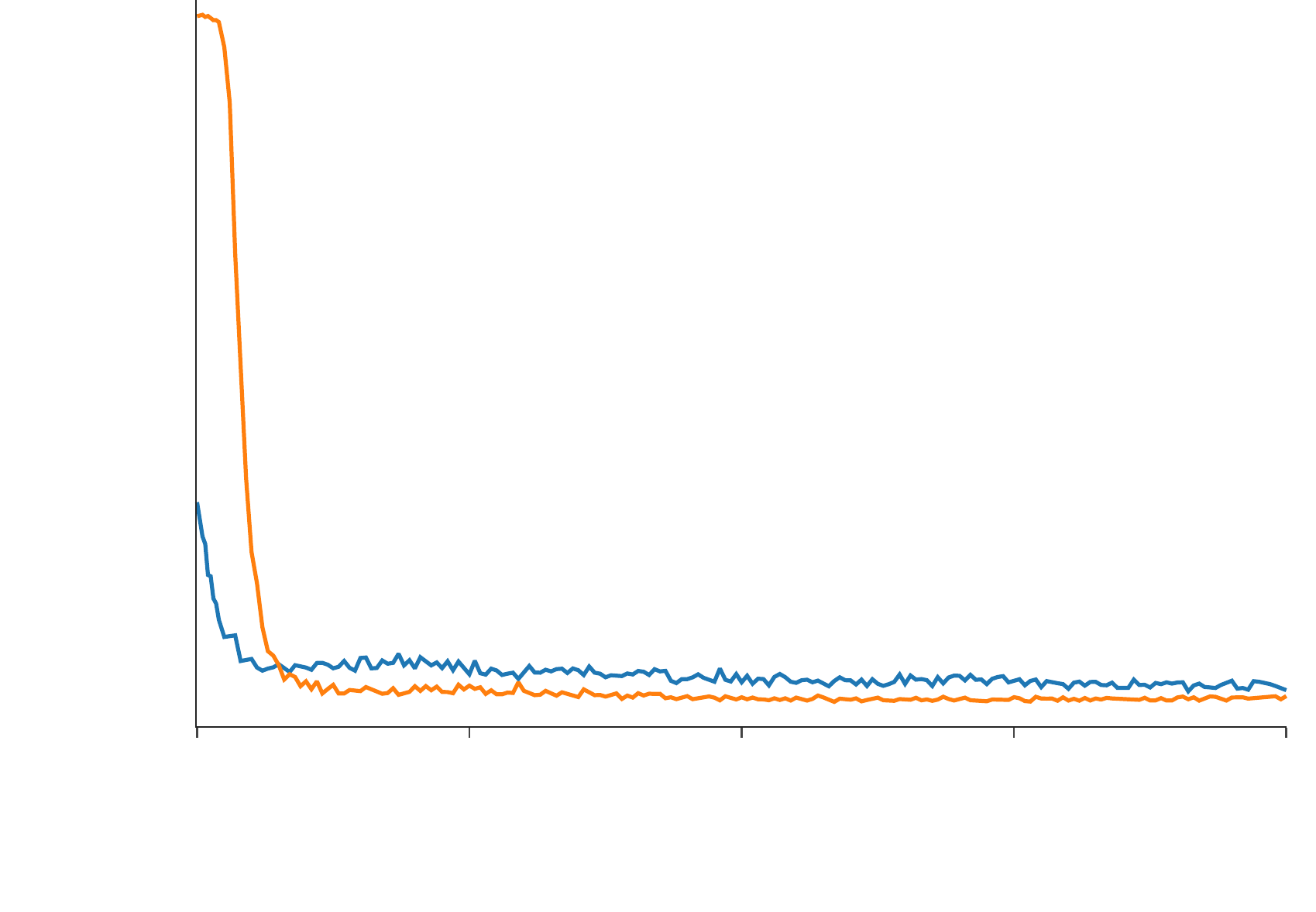
		\caption{Mean}
		\label{fig:irl_mean_distance}
	\end{subfigure}
	\begin{subfigure}[b]{0.49\columnwidth}
		\centering
		\def\svgwidth{\columnwidth}
		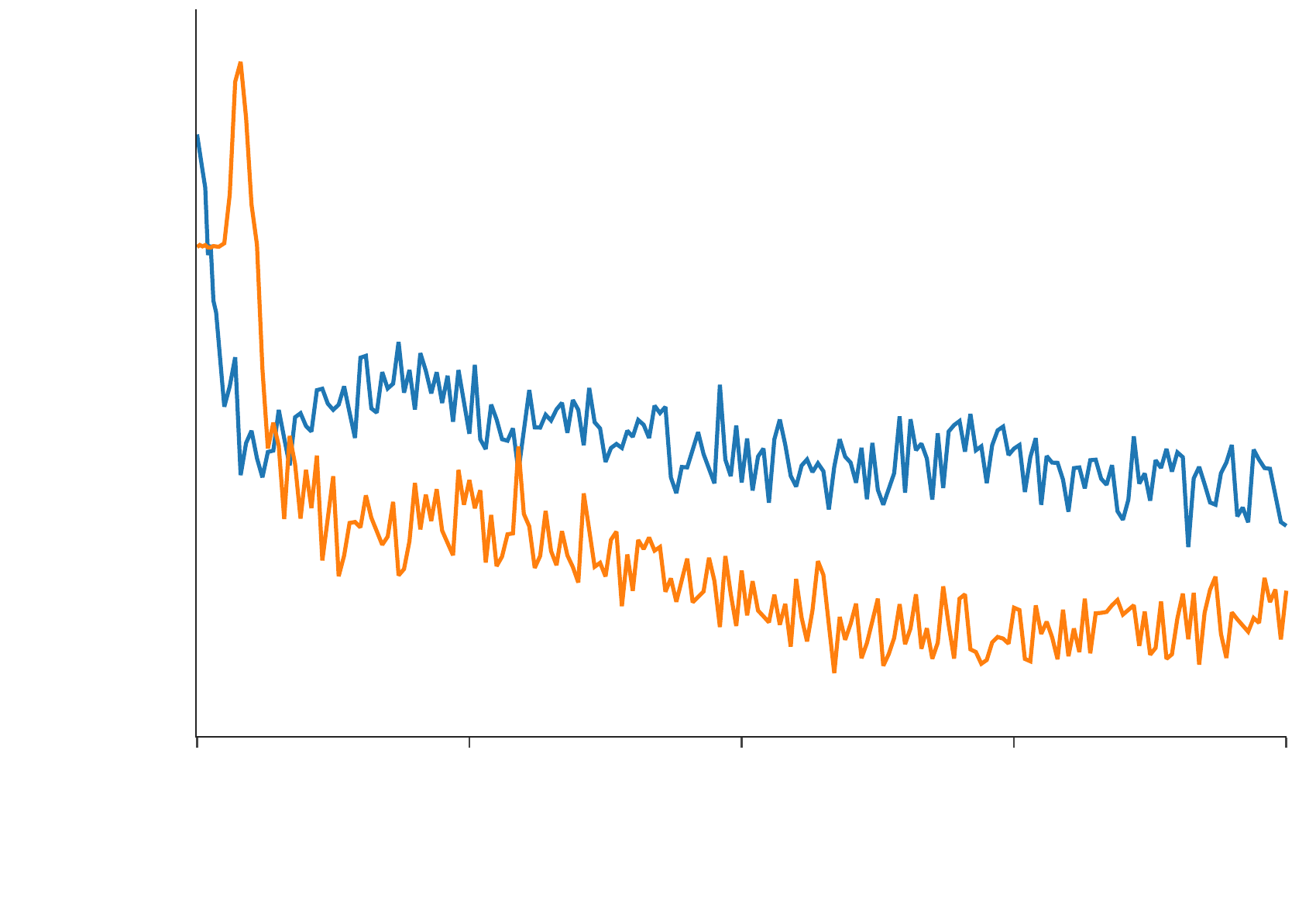
		\caption{Standard Deviation}
		\label{fig:irl_sd_distance}
	\end{subfigure}
	\caption{Distance Mean and Standard Deviation between $\trajectorySpace_\expert$ and $\trajectorySpace_\sample$: The Euclidean distance to the k-nearest neighbors ($k = 3$) expert trajectories throughout the training for the linear (blue) and nonlinear reward model (orange)}
	\label{fig:irl_mean_sd_distance}
\end{figure}

The absolute performance of the models in comparison to the manually tuned baseline is presented in Table~\ref{tab:irl_summary_sc01_sc06}.
The reward model of the baseline has been hand-tuned over numerous days through an iterative process of parameter modification and quantitative and qualitative analysis of the resulting trajectories.
It can be seen that both models perform well in all scenarios. The nonlinear model outperforms the manually tuned baseline in Sc02 and Sc06, as it does not generate any collisions or invalid trajectories. Further, the learned models manage to reach the desired velocity $\desiredVelocity{}$ in an additional \SI{46}{\percent} of the cases, while the desired lane is reached less frequently \SI{-5}{\percent}. Finally, both models yield a lower mean distance to the expert trajectories than the manually tuned baseline, while only the nonlinear model is consistently better.
A Euclidean distance metric is depicted in Fig.~\ref{fig:irl_mean_sd_distance}. As expected, the linear and the nonlinear model converge toward the expert trajectories. However, neither model converges completely but stalls at \SI{1.98}{\meter} and \SI{1.30}{\meter} for the linear and nonlinear, respectively. A possible remedy could be a reward model with a higher capacity.

\begin{figure*}
	\centering
	\begin{subfigure}[b]{\columnwidth}
		\centering
		\includegraphics[width=\columnwidth]{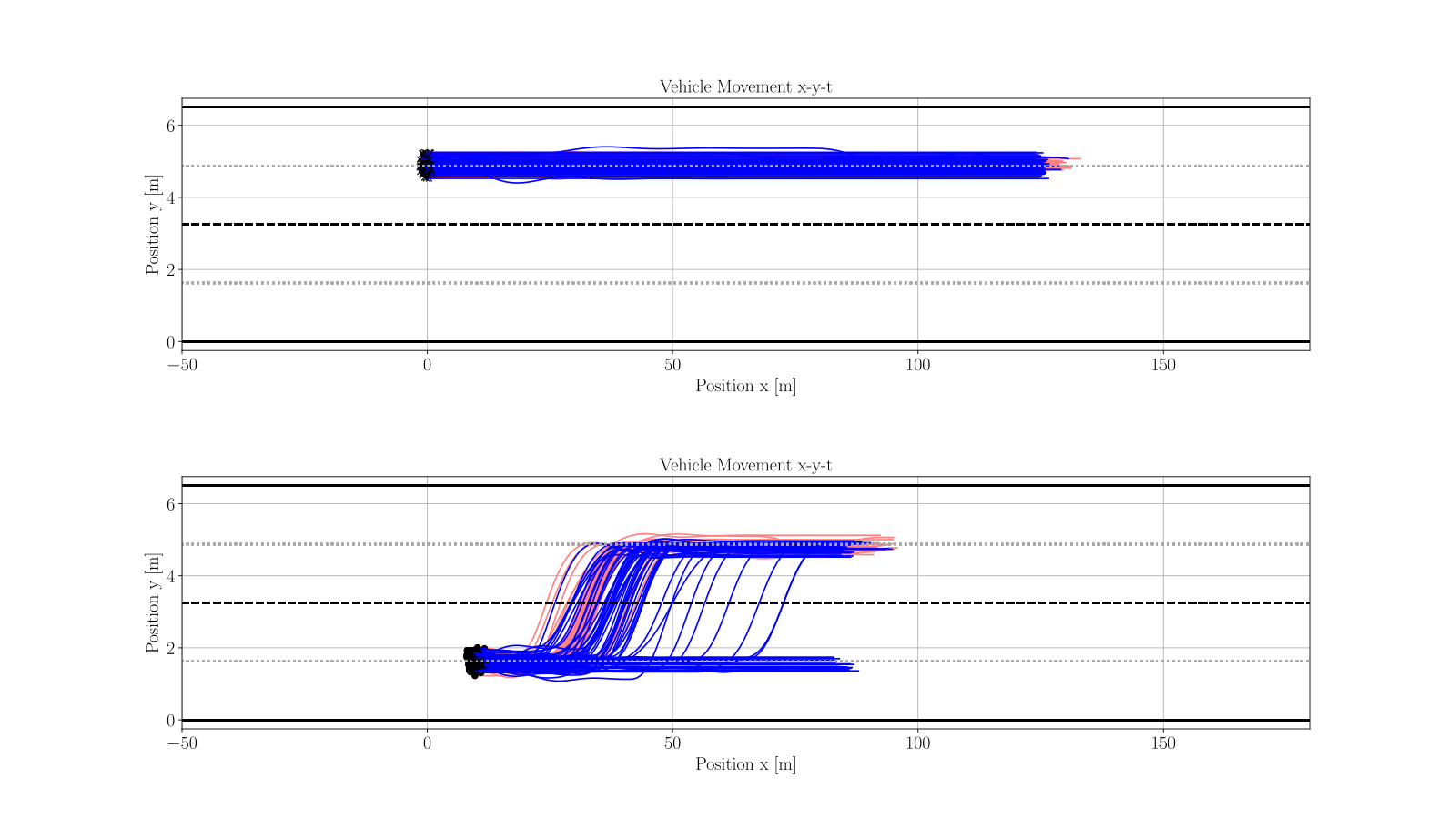}
		\caption*{(Sc01) Delaying merge due to approaching vehicle in desired lane}
		\label{fig:sc01}
	\end{subfigure}
	\hfill
	\begin{subfigure}[b]{\columnwidth}
		\centering
		\includegraphics[width=\columnwidth]{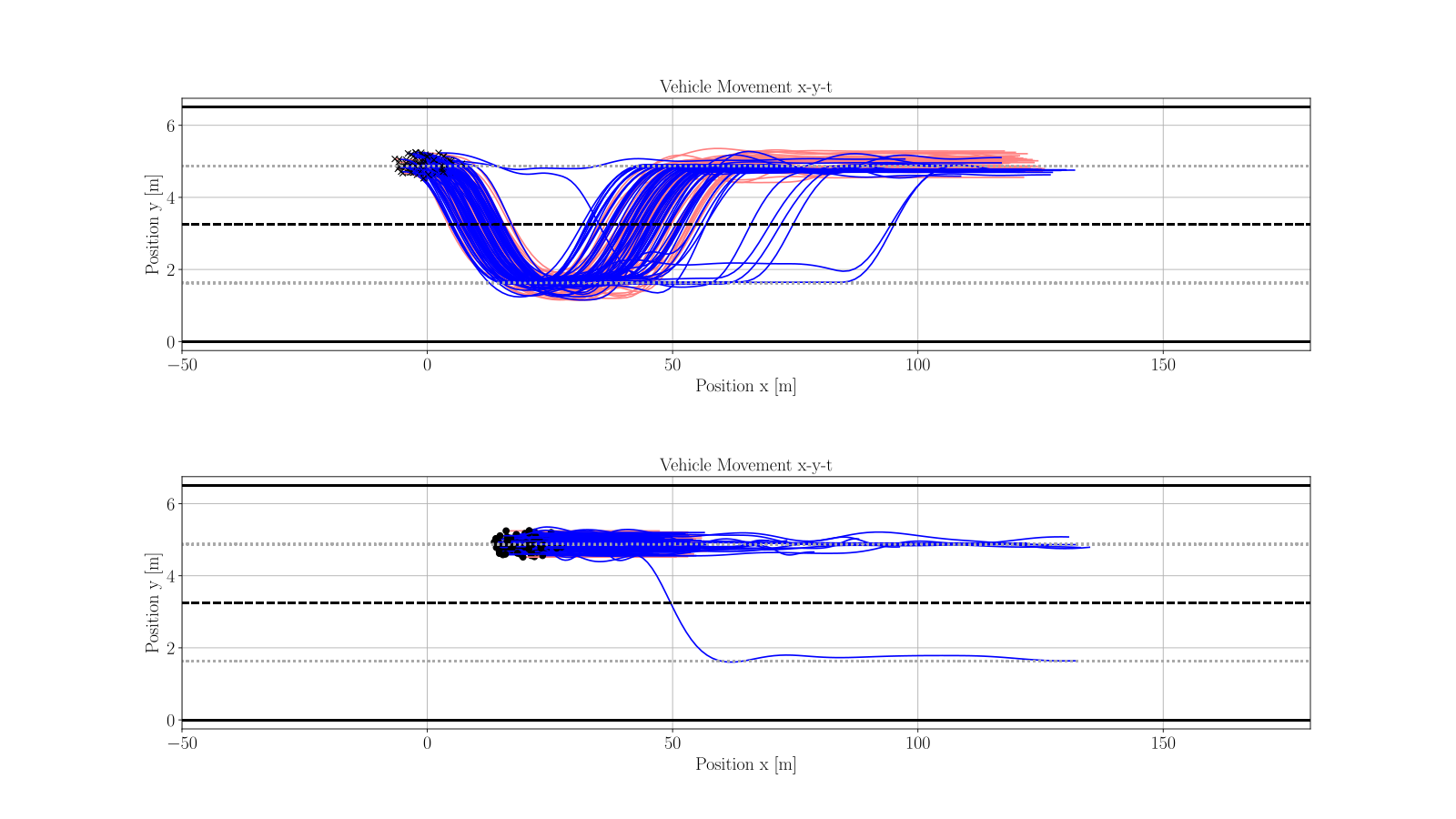}
		\caption*{(Sc02) Reacting to approaching vehicle}
		\label{fig:sc02}
	\end{subfigure}
	\hfill
	\begin{subfigure}[b]{\columnwidth}
		\centering
		\includegraphics[width=\columnwidth]{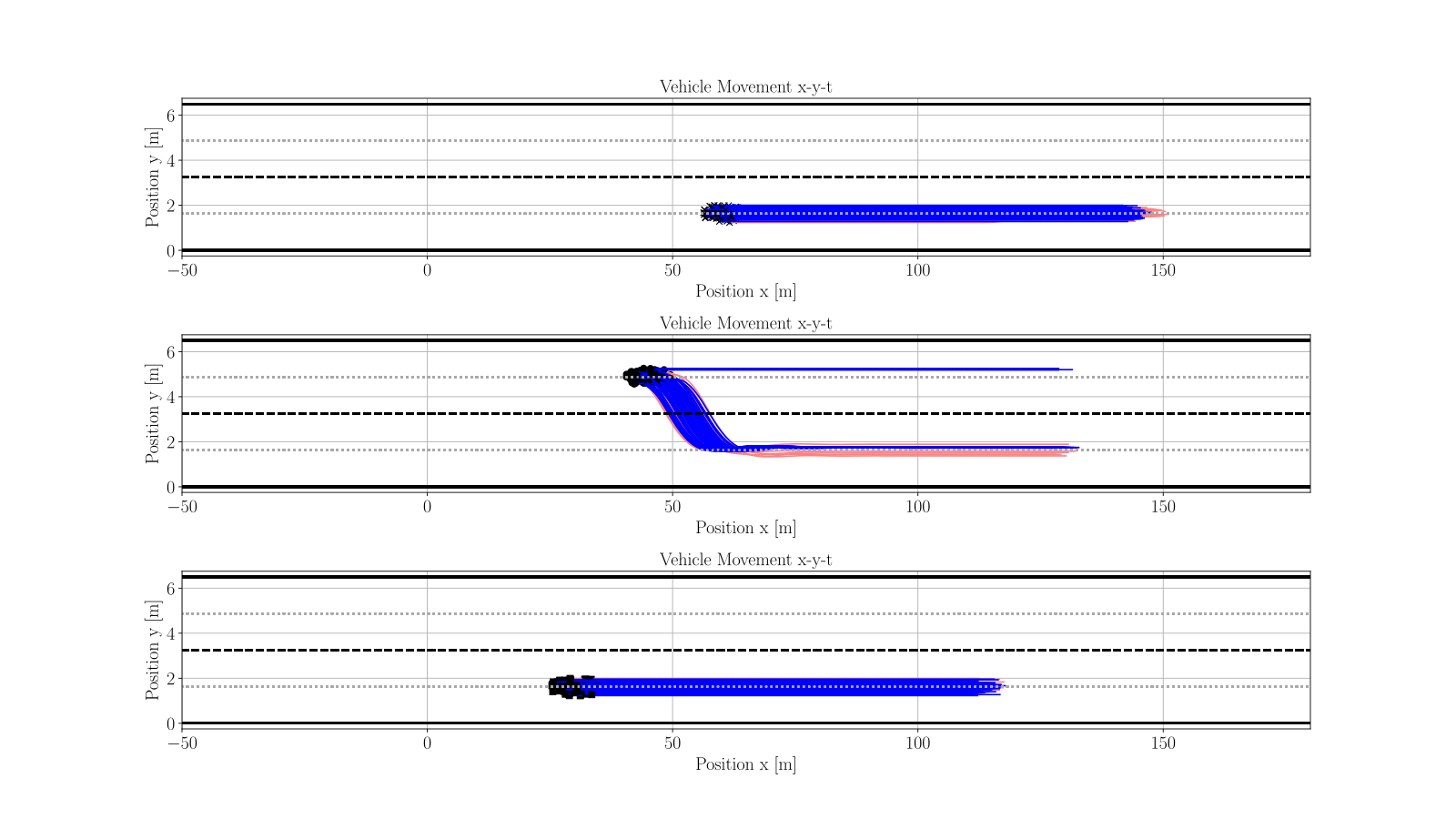}
		\caption*{(Sc03) Merging into moving traffic}
		\label{fig:sc03}
	\end{subfigure}
	\begin{subfigure}[b]{\columnwidth}
		\centering
		\includegraphics[width=\columnwidth]{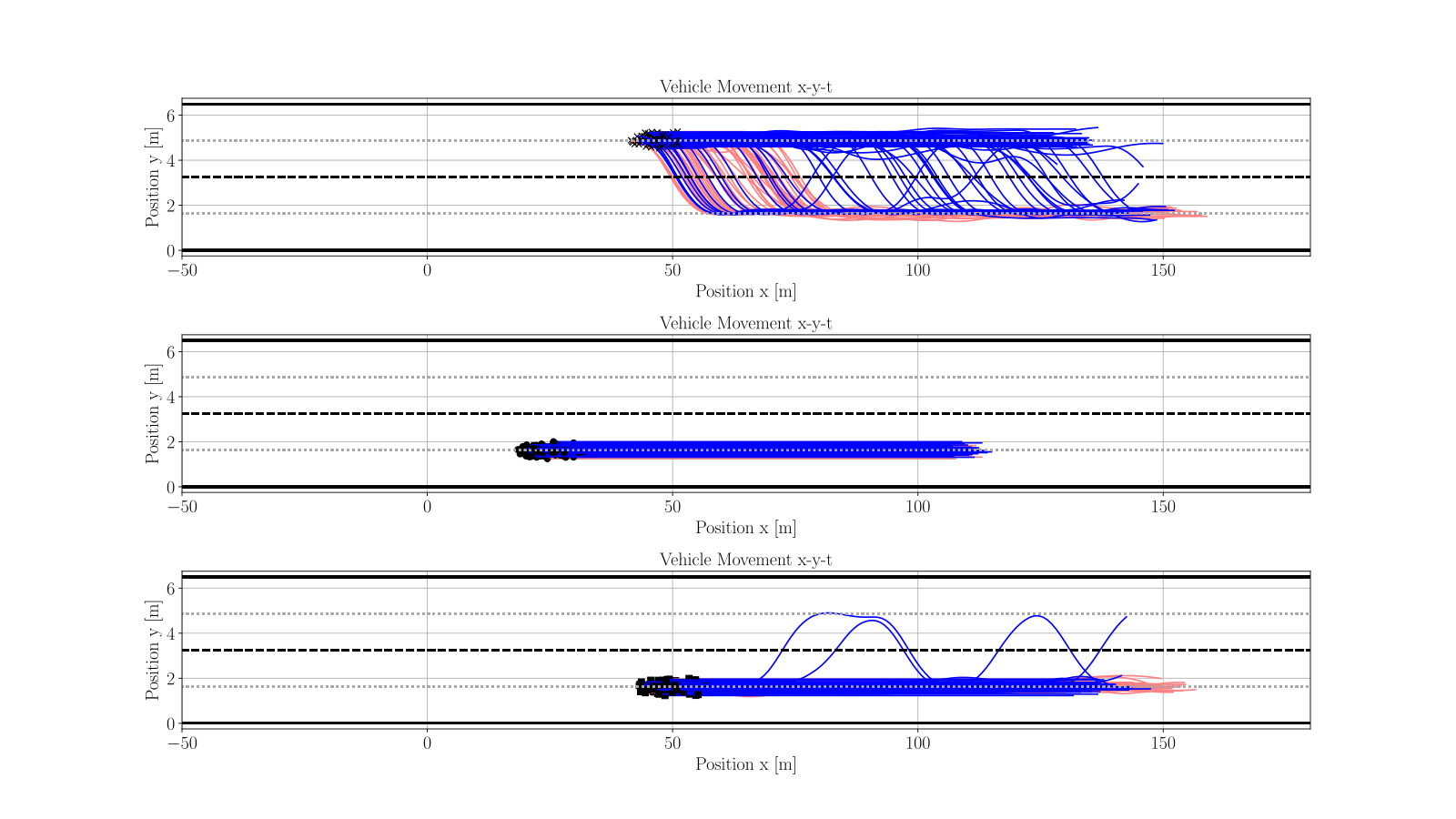}
		\caption*{(Sc04) Merging into moving traffic with prior longitudinal adjustment}
		\label{fig:sc04}
	\end{subfigure}
	\hfill
	\begin{subfigure}[b]{\columnwidth}
		\centering
		\includegraphics[width=\columnwidth]{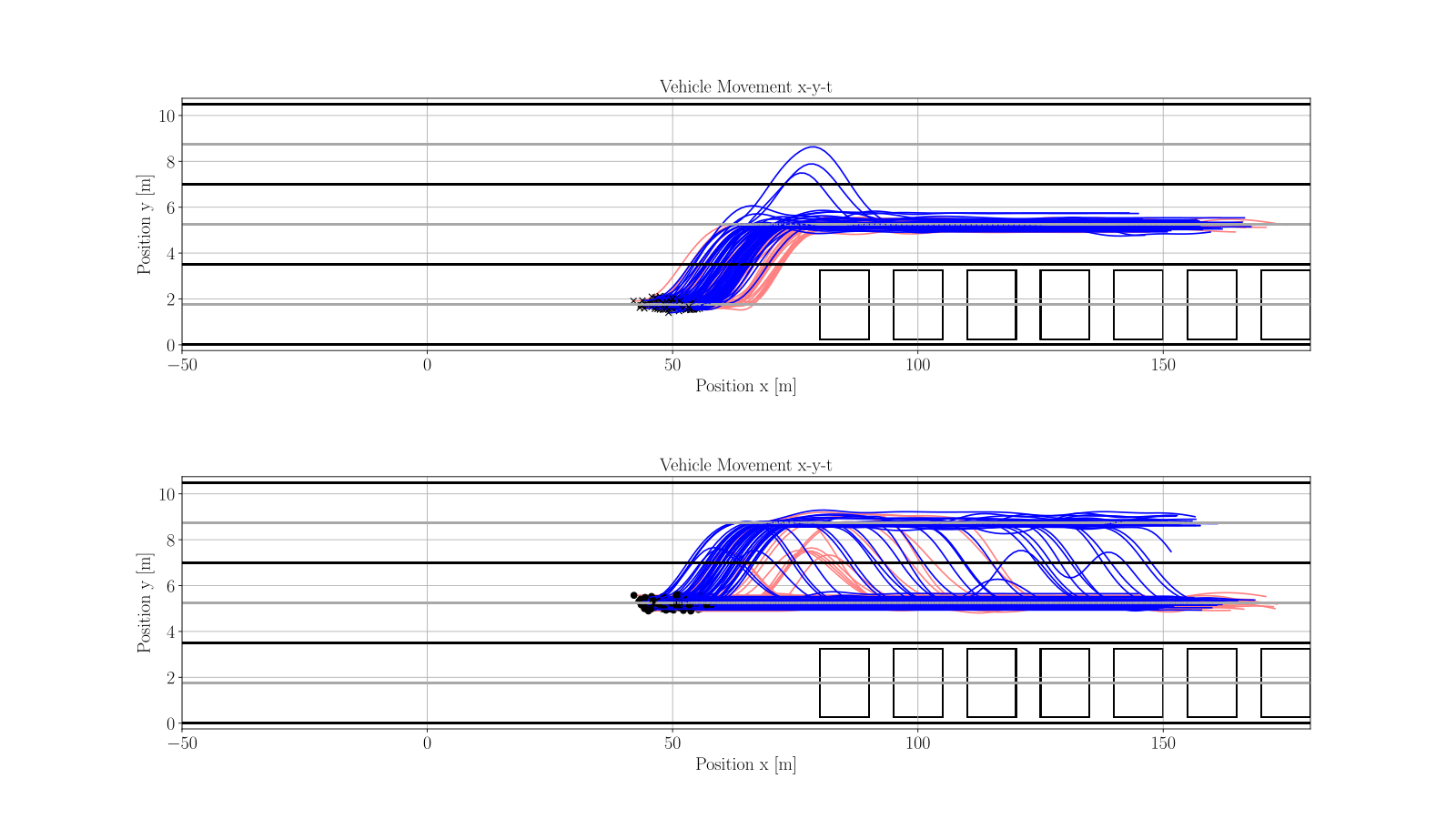}
		\caption*{(Sc05) Changing lane as other vehicle needs to merge onto lane}
		\label{fig:sc05}
	\end{subfigure}
	\hfill
	\begin{subfigure}[b]{\columnwidth}
		\centering
		\includegraphics[width=\columnwidth]{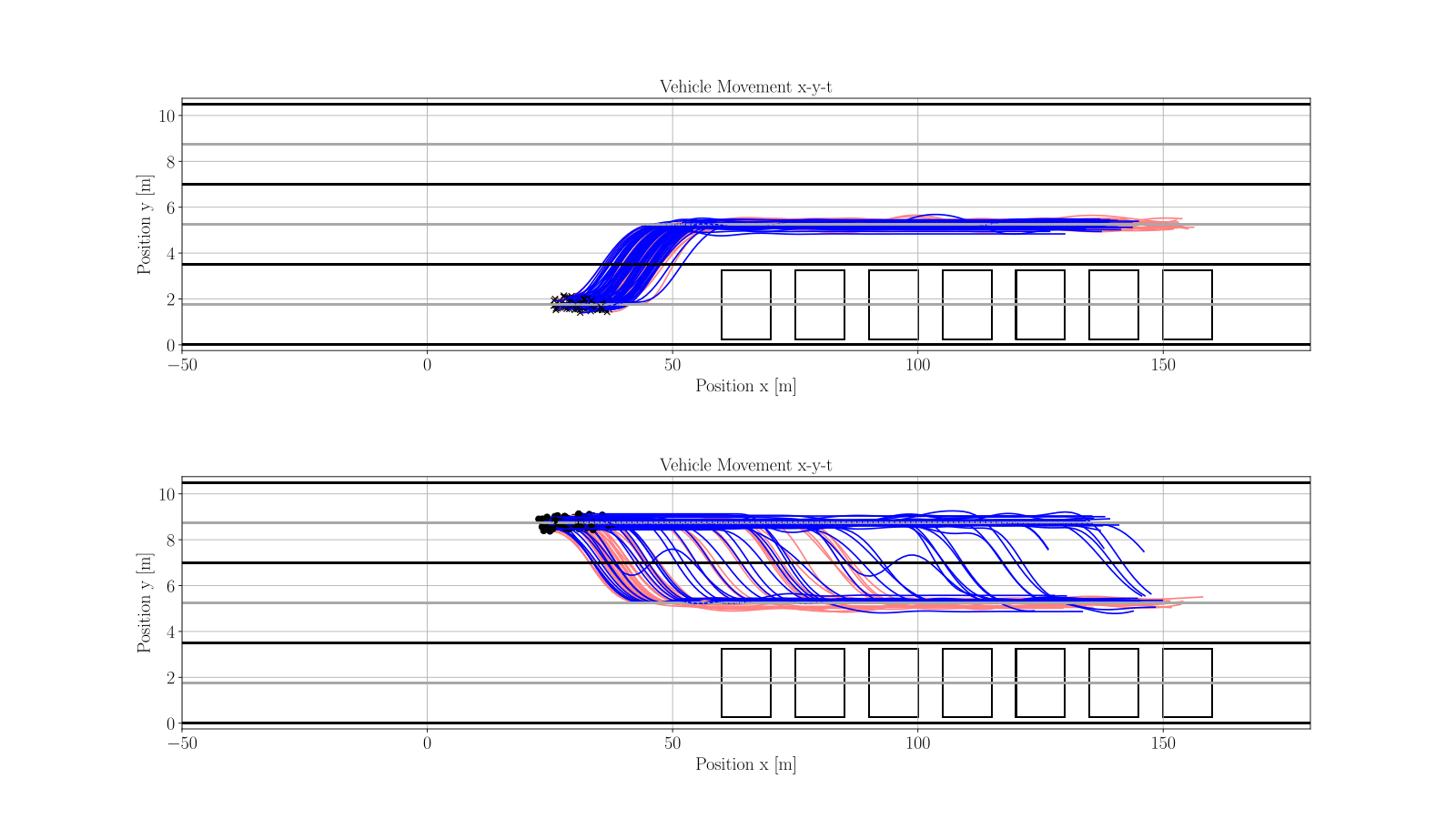}
		\caption*{(Sc06) Delaying lane change as other vehicle needs to merge first}
		\label{fig:sc06}
	\end{subfigure}
	\caption{Sample Trajectories from the Nonlinear Reward Model: Expert trajectories (red) that are used to learn the parameters of the reward model and the optimal trajectories (blue, after 2000 training steps of the nonlinear reward model)}
	\label{fig:sampled_trajectories}
\end{figure*}

The visual resemblance of the generated samples to the expert trajectories by the nonlinear reward model can be assessed in Fig.~\ref{fig:sampled_trajectories}.
Some of the trajectories that deviate significantly from the majority of the optimal trajectories could be the result of the inherently stochastic sampling-based trajectory planning algorithm.
\begin{table}
	\caption{Absolute Change in Performance: The performances of the linear and nonlinear reward models are compared with the manually tuned baseline on the scenarios. The columns collision, invalid, $\desiredLane{}$, and $\desiredVelocity{}$ denote the fraction of trajectories compared to the baseline that reaches that feature. Similarly, $\mu(d)$ and $\sigma(d)$ represent the mean and standard deviation of the Euclidean distance to the k-nearest neighbors ($k = 3$) expert trajectories compared to the baseline.}
	\label{tab:irl_summary_sc01_sc06}
	\setlength\tabcolsep{5pt}
	\begin{tabularx}{\columnwidth}{ l l r r r r r r}
		\toprule
		\textbf{Sc}			& \textbf{Model}& \textbf{collision}	& \textbf{invalid}		& \textbf{$\desiredLane{}$}		& \textbf{$\desiredVelocity{}$}		& \textbf{$\mu(d)$}		& \textbf{$\sigma(d)$}\\
		\toprule
		\multirow{2}{*}{Sc01}		& linear		& 0.00		& 0.00	& 0.06		& 0.03		& -0.02		& 0.06\\
									& nonlinear		& 0.00		& 0.00	& 0.07		& 0.01		& -0.04		& 0.03\\ \midrule
		\multirow{2}{*}{Sc02}		& linear		& 0.01		& 0.00	& 0.31		& 0.50		& -1.56		& 0.15\\
									& nonlinear		& 0.00		& -0.01	& 0.09		& 0.79		& -3.58		& -0.80\\ \midrule
		\multirow{2}{*}{Sc03}		& linear		& 0.00		& 0.00	& 0.02		& 0.61		& -1.14		& -0.25\\
									& nonlinear		& 0.00		& 0.00	& 0.03		& 0.60		& -1.22		& -0.35\\ \midrule
		\multirow{2}{*}{Sc04}		& linear		& 0.00		& 0.00	& -0.20		& 0.61		& -0.49		& 0.34\\
									& nonlinear		& 0.00		& 0.00	& -0.11		& 0.48		& -0.40		& 0.35\\ \midrule
		\multirow{2}{*}{Sc05}		& linear		& 0.05		& 0.01	& -0.25		& 0.51		& 1.54		& 0.27\\
									& nonlinear		& 0.00		& 0.00	& -0.17		& 0.41		& -0.38		& 0.27\\ \midrule
		\multirow{2}{*}{Sc06}		& linear		& -0.01		& 0.00	& -0.27		& 0.58		& -0.67		& 0.42\\
									& nonlinear		& -0.01		& 0.00	& -0.15		& 0.41		& -0.76		& 0.28\\ \midrule
		\multirow{2}{*}{Mean}		& linear		& 0.01		& 0.00	& -0.06		& 0.47		& -0.39		& 0.16\\
									& nonlinear		& 0.00		& 0.00	& -0.04		& 0.45		& -1.06		& -0.04\\ \bottomrule
	\end{tabularx}
\end{table}

\section{CONCLUSION}
This work combines Maximum Entropy Inverse Reinforcement Learning with Monte Carlo Tree Search to learn reward models for a cooperative multi-agent trajectory planning problem. The efficacy of the MCTS to generate (approximately) optimal samples for arbitrary reward models quickly in combination with the adjustment of the sampling distribution after gradient updates yield reward models that quickly converge towards the experts.

The results indicate that future trajectory planning algorithms that interact with humans in traffic might not need to rely on experts specifying features and parameters but on driving data from expert human drivers.

Thus, a possible research direction is the evaluation of nonlinear models with a higher capacity on a greater variety of scenarios, potentially even learning from raw data.



\section*{APPENDIX}
\subsection{Algorithms}
\begin{algorithm}
	\caption{Softmax Q-IRL}
	\label{alg:irl_softmax_q}
	\SetKwFunction{generateSamples}{generateSamples}
	\KwIn{$\trajectorySpace_\expert$}
	\KwOut{$\weights$}
	$\weights_0 \sim U[-1,1]$\;
	\For{$i \leftarrow 0$ \KwTo $M$}{
		$\trajectorySpace_\sample \leftarrow \varnothing$\;
		\For{$j \leftarrow 0$ \KwTo $N$}{
			$\trajectorySpace_\sample, \policySpace_\sample \leftarrow (\trajectorySpace_\sample, \policySpace_\sample) \cup$ \generateSamples{$\weights$}\;\label{alg:generate_samples}
		}
		$\widehat{\nabla_{\weights} \logLikelihood(\weights)} = \gradientReturnE - \frac{1}{|\trajectorySpace_\sample|} \sum_{\trajectory \in \trajectorySpace_\sample}\frac{e^{\return_{\weights}(\trajectory)}}{\policyTrajectory_\sample(\trajectory)\partitionFApprox}\gradientReturn$\;
		$\weights_{i+1} \leftarrow \weights_i + \alpha\widehat{\nabla_{\weights}\logLikelihood(\weights_i)}$\;
	}
	
	\KwRet{$\weights$}
\end{algorithm}
\begin{algorithm}
	\caption{Sampling of Trajectories and Policies}
	\label{alg:irl_data_sampling}
	\SetKwFunction{generateSamples}{generateSamples}
	\SetKwFunction{MCTSQEstimate}{MCTSQEstimate}
	\SetKwFunction{EnvironmentStep}{EnvironmentStep}
	\SetKwProg{Fn}{Function}{}{end}
	\Fn{\generateSamples{$\weights$}}{
		$\trajectorySpace\leftarrow \varnothing$; $\policySpace\leftarrow \varnothing$; $\policyTrajectory(\cdot) \leftarrow 1$\;
		\tcp{sample from the start state distribution}
		$\stateZero \sim d$\;
		\For{$t \leftarrow 0$ \KwTo $T-1$}{
			\tcp{estimate for each action explored at the root state}
			$\widehat{Q}(\stateT, \action_0), \dots , \widehat{Q}(\stateT, \action_m) \leftarrow$ \MCTSQEstimate{$\weights, \stateT$}\;
			$\policy[MCTS](\action | \stateT) \leftarrow \frac{e^{c{\widehat{Q}(\stateT, \action)}}}{\sum_{\action \in \actionSpace(\stateT)} e^{c{\widehat{Q}(\stateT, \action)}}}$\;
			\tcp{for each agent in the scenario}
			\For{$i \leftarrow 0$ \KwTo $|\agentSpace|$}{
				$\action_i \sim \policy[MCTS](\action | \stateT)$\;
				$\policyTrajectory(\trajectory_i) \leftarrow \policyTrajectory(\trajectory_i) \policy[MCTS](\action_i | \stateT)$\;
				$\trajectory_i \leftarrow \trajectory_i \cup (\stateT, \action_i)$\;
				\If{$t = T-1$}{
					$\trajectorySpace \leftarrow \trajectorySpace \cup \trajectory_i$\;
					$\policySpace \leftarrow \policySpace \cup \policyTrajectory(\trajectory_i)$\;}
			}
			$\stateT \leftarrow$ \EnvironmentStep{$\stateT, \actionZero, \dots, \action_m$}\;
		}
		\KwRet{$\trajectorySpace, \policySpace$}
	}
\end{algorithm}
\subsection{Importance Sampling}\label{sec:preliminiaries_importance_sampling}
Importance sampling allows to estimate a random variable $x$ following a distribution $p(x)$ by sampling from another distribution $q(x)$ \cite{Owen2013}.
\begin{equation}
    \begin{split}
	    \expectation_{X \sim p(x)}[X] & = \int_{\mathcal{D}} f(x) p(x) \mathrm{d} x\\
	    &=\int_{\mathcal{D}} \frac{f(x) p(x)}{q(x)} q(x) \dif x \\
	    &= \expectation_{X \sim q(x)}\left[X\frac{p(x)}{q(x)}\right]
    \end{split}
\end{equation}
\begin{equation}
	\widehat{\expectation}_{X \sim p(x)}[X] = \frac{1}{|\randomSet|} \sum_{x \in \randomSet} x \quad \randomSet \sim p(x)
\end{equation}
\begin{equation}
	\widehat{\expectation}_{X \sim q(x)}[X] = \frac{1}{|\randomSet|} \sum_{x \in \randomSet} x \frac{p(x)}{q(x)} \quad \randomSet \sim q(x)
\end{equation}
\subsection{Expectation of the Gradient of the Return}
\begin{equation}
	\label{eq:irl_expectation_gradient_return_derivation}
	\begin{split}
		\expectation_{\trajectory \sim \policyTrajectory_\expert(\trajectory)}\left[\gradientReturn\right] & = \int \gradientReturn \policyTrajectory_\expert(\trajectory) \dif\trajectory \\
		& = \int \gradientReturn \policyTrajectoryFull \dif\trajectory \\
		& = \int \gradientReturn \policyTrajectoryFull \frac{\policyTrajectory_\sample(\trajectory)}{\policyTrajectory_\sample(\trajectory)} \dif\trajectory \\
		& = \expectation_{\trajectory \sim \policyTrajectory_\sample(\trajectory)}\left[\frac{e^{\return_{\weights}(\trajectory)}}{\policyTrajectory_\sample(\trajectory)\partitionF}\gradientReturn\right]
	\end{split}
\end{equation}
\subsection{Approximation of the Partition Function}
\begin{equation}
	\label{eq:irl_partition_function_expectation}
	\begin{split}
		\partitionF & = \int e^{\return_{\weights}(\trajectory)} \dif\trajectory \\
		& = \int e^{\return_{\weights}(\trajectory)} \frac{\policyTrajectory_\sample(\trajectory)}{\policyTrajectory_\sample(\trajectory)}\dif\trajectory\\
		& = \expectation_{\trajectory \sim \policyTrajectory_\sample(\trajectory)} \left[\frac{e^{\return_{\weights}(\trajectory)}}{\policyTrajectory_\sample(\trajectory)}\right]\\
		&\approx \frac{1}{|\trajectorySpace_\sample|} \sum_{\trajectory \in \trajectorySpace_\sample} \frac{e^{\return_{\weights}(\trajectory)}}{\policyTrajectory_\sample(\trajectory)}
	\end{split}
\end{equation}

\section*{ACKNOWLEDGMENT}
We wish to thank the German Research Foundation (DFG) for funding the project Cooperatively Interacting Automobiles (CoInCar), within which the research leading to this contribution was conducted. The information and views presented in this publication are solely the ones expressed by the authors.
\vfill
\bibliographystyle{IEEEtran}
\bibliography{library}

\end{document}